\documentclass[10pt,twocolumn,letterpaper]{article}

\usepackage[pagenumbers]{wacv} 

\usepackage{graphicx}
\usepackage{amsmath}
\usepackage{amssymb}
\usepackage{booktabs}
\usepackage[accsupp]{axessibility}
\usepackage[inline]{enumitem}
\usepackage{tabularx}
\usepackage[absolute,overlay]{textpos}

\newcolumntype{d}{X}
\newcolumntype{s}{>{\hsize=.35\hsize}r}

\usepackage{textcomp}
\usepackage{siunitx}

\usepackage[pagebackref,breaklinks,colorlinks]{hyperref}

\usepackage[capitalize]{cleveref}
\crefname{section}{Sec.}{Secs.}
\Crefname{section}{Section}{Sections}
\Crefname{table}{Table}{Tables}
\crefname{table}{Tab.}{Tabs.}

\def\wacvPaperID{435} 
\def\confName{WACV}
\def\confYear{2024}

\begin{document}

\begin{textblock*}{\textwidth}(20mm,10mm) 
\noindent
\small \copyright 2023 IEEE. Personal use of this material is permitted. Permission from IEEE must be obtained for all other uses, in any current or future media, including reprinting/republishing this material for advertising or promotional purposes, creating new collective works, for resale or redistribution to servers or lists, or reuse of any copyrighted component of this work in other works.
\end{textblock*}

\title{ISAR: A Benchmark for Single- and Few-Shot Object \textbf{I}nstance \textbf{S}egmentation \textbf{a}nd \textbf{R}e-Identification}

\author{Nicolas Gorlo\\
{\tt\small gorlon@ethz.ch}\and
Kenneth Blomqvist\\
{\tt\small kblomqvist@mavt.ethz.ch}\and
Francesco Milano\\
{\tt\small francesco.milano@mavt.ethz.ch}\and
Roland Siegwart\\
{\tt\small rsiegwart@ethz.ch}\\
Autonomous Systems Lab, ETH Z\"urich
}

\maketitle
\thispagestyle{empty}

\newcommand{\blue}[1]{\textcolor{blue}{#1}}
\newcommand{\todo}[1]{\blue{\textbf{(TODO: #1)}}}
\newcommand{\rev}[1]{{#1}}
\newcommand{\red}[1]{\textcolor{red}{#1}}
\newcommand{\FM}[1]{\blue{\textbf{(FM: #1)}}}
\newcommand{\KB}[1]{\blue{\textbf{(KB: #1)}}}

\begin{abstract}
Most object-level mapping systems in use today make use of an upstream learned object instance segmentation model. If we want to teach them about a new object or segmentation class, we need to build a large dataset and retrain the system. To build spatial AI systems that can quickly be taught about new objects, we need to effectively solve the problem of single-shot object detection, instance segmentation and re-identification. So far there is neither a method fulfilling all of these requirements in unison nor a benchmark that could be used to test such a method. Addressing this, we propose \textbf{ISAR}, a benchmark and baseline method for single- and few-shot object \textbf{I}nstance \textbf{S}egmentation \textbf{A}nd \textbf{R}e-identification, \rev{in an effort} to accelerate the development of algorithms that can robustly detect, segment, and re-identify objects from a single or a few sparse training examples. We provide a semi-synthetic dataset of video sequences with ground-truth semantic annotations, a standardized evaluation pipeline, and a baseline method. Our benchmark aligns with the emerging research trend of unifying Multi-Object Tracking, Video Object Segmentation, and Re-identification.

\end{abstract}

\vspace{-10pt}
\section{Introduction}
\label{sec:introduction}

Object-level scene understanding is fundamental to the development of robust and effective spatial AI systems. The ability for an AI system to accurately perceive objects within a scene is crucial to many applications in robotics, augmented reality, navigation, and autonomous vehicles. 

Working towards the goal of general-purpose, spatial AI systems, the research community has become increasingly interested in developing object-level mapping pipelines~\cite{runz2017co, xiao2019dynamic, cui2019sof, volumetric_instance-aware_semantic_mapping, slam++, vmap, Kimera, Fusionpp, kundu2022panoptic, strecke2019fusion, xu2019mid, zhang2020vdo, bescos2021dynaslam, liu2019global} for use in robotics, augmented reality and other spatial AI applications. All of these pipelines assume known object models~\cite{slam++} or an upstream object segmentation pipeline, either to assign a semantic class to static parts of the scene~\cite{volumetric_instance-aware_semantic_mapping, Kimera}, or to deal with dynamic objects separately from the static background mapping and reconstruction~\cite{runz2017co, xiao2019dynamic, cui2019sof, strecke2019fusion, xu2019mid, zhang2020vdo, bescos2021dynaslam, vmap, kundu2022panoptic, liu2019global}. Additionally, systems have been proposed which incrementally build databases of objects and their geometry to help with object manipulation \cite{incremental_object_database, lu2022online}. Such systems also rely on object instance segmentation. 
Crucially, this object-level segmentation is currently produced using learned object segmentation models such as Mask R-CNN~\cite{maskrcnn}, Sharp-Mask~\cite{pinheiro2016learning}, PSPNet~\cite{zhao2017pyramid} or Detic~\cite{detic}. Such learned object segmentation models require classes to be known at training and dataset building time. Fine-tuning them to handle a new class requires annotating thousands of diverse examples and many expensive optimization iterations. To build general-purpose robots and spatial AI systems that can deal with arbitrary objects, one needs to be able to teach them about new objects at run-time. Teaching the spatial AI system and adding new objects to an object tracking and segmentation pipeline should be as easy as selecting the new objects through a user interface. Effectively, this means solving the problem of single- or few-shot object instance segmentation and re-identification. 

So far, the computer vision community has studied the problems of few-shot semantic segmentation, Video Object Segmentation (VOS) and Re-identification (re-ID) individually. While tremendous progress has been made on all of these tasks independently, we can still not teach our spatial AI systems effectively about new objects. Few-shot semantic segmentation systems use dense object segmentation masks, which are expensive to obtain, and they do not make use of the temporal structure in the video data that spatial AI systems inherently operate on. VOS methods perform extremely well, and make full use of video data, but require an initial dense segmentation mask. Further, they do not deal with re-identifying the objects across different \emph{scene contexts}, \rev{which we define as the surrounding environment in which an object is recorded, including the pose of the objects in the environment}. As a result, the in-the-wild usability of VOS methods for the previously mentioned tasks is limited.

In summary, existing methods for object instance segmentation and re-identification either do not use the temporal structure of video data, rely on initial dense segmentation masks and/or struggle to re-identify objects across different scenes. Consequently, there is an apparent gap which could be filled by combining few-shot semantic segmentation, VOS and re-ID methods into one approach.

To address this gap, we propose \textbf{ISAR}, a benchmark and baseline method for single- and few-shot object \textbf{I}nstance \textbf{S}egmentation \textbf{A}nd \textbf{R}e-identification. Our benchmark is designed to accelerate research progress towards robust algorithms that will enable teaching spatial AI systems through a single or a few sparse training examples. We aim to unify few-shot semantic segmentation, Video Object Segmentation (VOS) and Re-identification (re-ID) -- traditionally mostly separately studied research topics. This reflects the emerging trend in recent years towards combining these topics~\cite{mot_lit_review}. As our goal is to build spatial AI systems that can deal with objects in any configuration, including moving objects, we aim to make it hard to rely purely on spatial information as the primary means of object description. Instead, we advocate the development of vision-based object-centric methods, which do not rely on the context in which the object is first perceived. Through this approach, we strive to advance the representations of object instances, \rev{making them more detailed and distinguishable, which in turn may} enable more robust re-identification and object-level mapping of dynamic environments. 
\rev{Our} semi-synthetic benchmark dataset is recorded using the Habitat AI Simulator~\cite{habitat, habitat2} with scenes of Replica~\cite{replica} and the Habitat-Matterport 3D research dataset~\cite{hm3d}, using objects from the YCB dataset~\cite{ycb}.
The dataset and evaluation code are available at \url{https://nicogorlo.github.io/isar_wacv24/}.

To \rev{summarize}, our contributions are the following:
\begin{enumerate*}
    \item A semi-synthetic dataset of video sequences with high-quality ground-truth semantic annotations;
    \item A standardized evaluation pipeline to measure the performance of different methods on this task against each other;
    \item A baseline method to segment and re-identify objects
\end{enumerate*}.
\section{Background}

In this section, we cover some of the work done on related problems, and show that none of these fully address the needs of modern object-centric spatial AI systems.

\subsection{Video Object Segmentation}
The primary goal of Video Object Segmentation (VOS) is to segment object instances in a video sequence. 
VOS has been categorized into four major subcategories: Semi-Supervised, Unsupervised, Referring and Interactive VOS.

\textbf{Semi-Supervised VOS} requires propagating a mask, provided in the first frame, to subsequent frames in a video. Facilitated by various benchmark datasets~\cite{youtube_vos, davis, davis2017, davis2018, davis2019, vot2020}, large steps in performance have been achieved~\cite{video_object_segmentation_with_reid, osvos, kmn, stm}. 
In particular, recent methods~\cite{xmem, qdmn, wang_look_2022} have shown very strong performance on these benchmarks. However, their reliance on a costly annotation in the initial frame and their dependence on the object being salient and staying in the same context reduces their in-the-wild applicability for spatial AI systems, as the initial annotation is hard to provide in an online setting and objects are not guaranteed to be \emph{salient} \rev{(\ie, prominent in the video frame)}.

In \textbf{Unsupervised VOS} the goal is to segment the salient objects in a video clip without any labeling cues or training. Using the same benchmark datasets as Semi-Supervised VOS and some specialized ones~\cite{DAVSOD, SegTrackv2, ViSal, FBMS}, equally impressive steps in performance of methods have been achieved~\cite{SegFlow, COSNet, 3d_conv_unsupervised}.
However, unsupervised VOS methods \rev{by definition} rely on the fact that the objects are salient and the overall scene context does not change. Consequently, it is not applicable when these assumptions are not met. \rev{However, in in-the-wild scenarios, there no guarantees that the assumptions are met.}\\
In \textbf{Video Instance Segmentation} (VIS), the goal is to segment all instances in a video from a set of object categories.
YouTube-VIS~\cite{youtube_vis}, the most popular benchmark for VIS is built on the YouTube-VOS~\cite{youtube_vos} benchmark dataset and therefore suffers from the same problem of few re-appearing and non-salient objects \rev{as YouTube-VOS}. Further, it is constrained to the predefined object categories.

The novel task of \textbf{Referring VOS} replaces the segmentation in the initial frame of Semi-Supervised VOS with a language prompt. The benchmarks Youtube-VOS~\cite{urvos} and DAVIS-2017~\cite{vos_language_referring_DAVIS17} have been extended with natural language prompts describing the objects. A number of methods have already achieved promising results on this task~\cite{urvos, botach_end--end_2022, wu_multi-level_2022, wu_language_2022, yofo, vos_language_referring_DAVIS17}.

In \textbf{Interactive VOS}~\cite{davis2018, davis2019} the initial full mask of the semi-supervised scenario is replaced with interactive user inputs, given as scribbles, to refine the video object segmentation throughout the video. The interactive user input is provided in up to $8$ rounds for the frame with the worst prediction among candidate frames. The strong performance of state-of-the-art~\cite{scribblebox, mivos} methods on this task shows that one does not have to rely on expensive mask annotations like Semi-Supervised VOS.

While many benchmarks have been created for different types of VOS~\cite{youtube_vos, davis, davis2017, davis2018, davis2019, youtube_vis, DAVSOD, SegTrackv2, ViSal, FBMS, MOSE, qi2022occluded}, these mostly focus on segmenting salient objects, which rarely move out of frame or reappear in a different context. In addition, recently it was shown on MOSE~\cite{MOSE} and OVIS~\cite{qi2022occluded}, datasets for VOS and VIS in complex scenes that state-of-the-art VOS and VIS methods, achieving near perfect scores on DAVIS 2016~\cite{davis}, struggle with heavy occlusion and more complex scenes. This strengthens the case that current VOS and VIS benchmark datasets do not properly mimic an in-the-wild scenario. However, even in the MOSE and OVIS benchmark datasets, the scene context mostly stays the same. Therefore, methods developed for these tasks are not fit to deal with reappearing, dynamic objects or for identifying previously seen objects in a new context.

Our benchmark, separating annotated and annotation-free scenes into distinct scene contexts, goes beyond the current scope of tasks in the field. This benchmark is designed to fuel the advancement of methods that combine VOS with re-identification across varying scene contexts. Instead of providing full segmentation masks, we provide point- and bounding box annotations as hints. This maintains some of the advantages of Interactive VOS, preventing methods from relying on costly segmentation.

\subsection{Vehicle and Person Re-identification}
\label{sec:re-id}
Instance re-identification has been mainly explored within vehicle and person re-identification, as well as for face recognition. Facilitated by datasets for person~\cite{mars, person_reid_bench, duke_mtmc, viper, prw, xiao2017joint, airport} and vehicle~\cite{stanford_cars, comp_cars, vehicleID, cityflow, veri_wild} re-identification, impressive accuracy has been achieved in distinguishing and re-identifying vehicles and people, despite few differences among the instances that need to be distinguished from one another. State-of-the-art methods~\cite{wang_spatial-temporal_2018, wieczorek_unreasonable_2021, fu_unsupervised_2021, zhu_viewpoint-aware_2019, ni_flipreid_2021, vreid_attributenet, vreid_groupsensitive, vreid_tripletembedding, vreid_vehiclenet} now effectively solve the task with little error on the most popular datasets.

However, direct application of these methods to other object classes is infeasible, as the methods rely on training on large datasets of the same class. These challenging circumstances demand the development of techniques that can perform efficiently even with scarce data or no prior class information. Further, these methods only tackle re-identification and not object instance segmentation from videos.

\rev{As we deal with re-identifying objects of any class rather than specific classes, our definition of re-ID differs from these standard re-ID frameworks in that we do provide single- or few-shot labels. In standard re-ID frameworks, no labels are provided, but rather all objects of a specific class need to be re-identified.}

\subsection{Few-shot Semantic Segmentation}

Few-shot semantic segmentation (FSS) \cite{shaban2017one, dong2018few, tian2020prior, fan2020fgn, fan2022self, peng2023hierarchical} methods predict pixelwise masks for novel classes given a few annotations. Most methods are based on metric learning \cite{dong2018few}. Later works have introduced support query features and attention mechanisms \cite{siam2020weakly, liu2020dynamic, yang2020brinet}, others have introduced fine-tuning \cite{zhu2021self}, memory modules \cite{wu2021learning}, and learned classifiers \cite{lu2021simpler}. 

FSS methods are typically evaluated on the \hbox{PASCAL-5\textsuperscript{i}} \cite{everingham2010pascal, shaban2017one} and the COCO-20 \cite{lin2014microsoft} datasets, which hold out a certain number of classes that are used for few-shot evaluation. In these datasets, the training set actually contains some instances of the test classes already, but those are not labeled. \rev{FFS methods} focus on simply segmenting the object class and do not detect or re-identify the same instances of an object. In the test phase, a full segmentation mask is provided, instead of a sparse prompt like a bounding box or pixel coordinate. They also operate on individual images instead of video, which rules out the use of methods which leverage temporal structure in the data. For an object-level SLAM type application, figuring out how to rely on sparser expert labels and how to make the best use of video data is necessary.





\section{Problem Formalization}

\label{sec:problem_formalization}
In the following, we denote with $\mathcal{D}^{\text{train}}$ and $\mathcal{D}^{\rev{\text{eval}}}$ respectively an annotated training \rev{set} and an evaluation \rev{set}, both of which we assume to consist of one or more ordered image sequences. Let the train sequences be indexed with \rev{$i \in \{1, \dots, \Omega^{\text{train}}\}$} and the evaluation sequences be indexed with \rev{$j \in \{1, \dots, \Omega^{\rev{\text{eval}}}\}$}.
In particular, we define a training sequence $\mathcal{I}^{\text{train}}_i = \{\rev{\rev{\mathbf{I}}}^{\text{train}}_1, \rev{\mathbf{I}}^{\text{train}}_2, ..., \rev{\mathbf{I}}^{\text{train}}_{N_i}\} \in \mathcal{D}^{\text{train}}$ as a sequence of $N_i$ posed RGB or RGB-D images \hbox{$\rev{\mathbf{I}}^{\text{train}}_r = (f_r, p_r)$} (\ie, tuples of image $f_r$ and $6$-DoF pose $p_r$ in the world coordinate frame), \rev{with $r \in \{1, \dots, N_i\}$}, and an evaluation sequence \hbox{$\mathcal{I}^{\rev{\text{eval}}}_j = \{\rev{\mathbf{I}}^{\rev{\text{eval}}}_1, \rev{\mathbf{I}}^{\rev{\text{eval}}}_2, ..., \rev{\mathbf{I}}^{\rev{\text{eval}}}_{M_j}\} \in \mathcal{D}^{\rev{\text{eval}}}$} as a sequence of $M_j$ posed RGB or RGB-D images.
Each training sequence $\mathcal{I}^{\text{train}}_i$ contains $K_i$ annotated objects. Each annotation consists of the image index in which the annotation occurs, a bounding box given by the 2D image coordinates of the top left and bottom right corners, and a point that is part of the object. Let $\mathcal{A}^{\text{train}}_i$ denote the set of all annotations for sequence $\mathcal{I}^{\text{train}}_i$.
For the train data annotation, there are two scenarios: a single-shot scenario and a multi-shot scenario. In the single-shot scenario, there is only one train sequence ($|\mathcal{D}^{\text{train}}| = 1$) and there is only one annotation per object. In the multi-shot scenario, there can be both more than one train sequence and more than one annotation per sequence.
The evaluation sequences only contain ground-truth mask annotations for evaluation.


For each object $O_k$ we seek to learn a mapping
\begin{align}
\label{eq:mapping}
\begin{split}
    F_k:~& \{\mathcal{I}^{\text{train}}_i \}_{i\le \rev{\Omega}^{\text{train}}}\times\{\mathcal{A}^{\text{train}}_i \}_{i \le \rev{\Omega}^{\text{train}}}\times\\
    &\{\rev{\mathbf{I}}^{\rev{\text{eval}}}_j(0), ..., \rev{\mathbf{I}}^{\rev{\text{eval}}}_j(t^\star)\} \in \mathcal{I}^{\rev{\text{eval}}}_j\\
    & \mapsto \rev{\mathbf{M}}_{k,j}^{\rev{\text{eval}}}(t^\star),
\end{split}
\end{align}
such that the discrepancy between $\rev{\mathbf{M}}_{k,j}^{\rev{\text{eval}}}(t^\star)$ and $\rev{\mathbf{M}}_{k,j,gt}^{\rev{\text{eval}}}(t^\star)$ is minimized for all frames in each evaluation sequence $\mathcal{I}^{\rev{\text{eval}}}_j$ and all objects $O_k$, $k \le K$.
Here $\rev{\mathbf{M}}_{k,j}^{\rev{\text{eval}}}(t^\star)$ denotes the predicted binary instance segmentation mask of object $O_k$ in the sequence $\mathcal{I}^{\rev{\text{eval}}}_j$ at time $t^\star$ and $\rev{\mathbf{M}}_{k,j,gt}^{\rev{\text{eval}}}(t^\star)$ its corresponding ground-truth annotation.
Therefore, the goal is to sequentially produce pixelwise masks on the evaluation sequences, given the training sequence and the annotations.

The desired properties of the mappings $F_k$ include:
\begin{enumerate*}
    \item \textit{Consistency with annotations}: For every object instance $O_k$, the mapping $F_k$, if applied to a train sequence $\mathcal{I}^{\text{train}}_i$ should produce segmentation masks that are consistent with the provided annotations.
    \item \textit{Temporal consistency}: Given the temporal nature of the input video sequences, the mapping $F_k$ should produce segmentation masks that are temporally consistent. In other words, for each object instance $O_k$, the predicted instance segmentation masks across different frames in a sequence should correspond to the same object.
    \item \textit{Spatial Consistency}: Both the predicted segmentation masks and its boundary should match the ground-truth segmentation mask.
    \item \textit{Generalization to different scenes}: The mappings $F_k$ should be able to generalize to entirely different scene contexts in the evaluation data and should therefore be able to produce coherent segmentation masks in all the sequences $\{\mathcal{I}^{\rev{\text{eval}}}_j\}_{\rev{j \in \{1, \dots, \Omega^{\rev{\text{eval}}}\}}}$ regardless of the scene context that the sequences may be recorded in.
\end{enumerate*}



We measure adherence to these properties with evaluation metrics introduced in~\Cref{sec:evaluation}.
The overall objective of this task is to derive such mappings $F_k$ that fulfill the desired properties. 
Importantly, the entire train sequence and the associated sparse annotation data can be used to form object representations that enable re-identification of the objects in a different scene context. An example of such object representations is outlined in~\Cref{sec:method_pipeline}.

For in-the-wild applicability, methods tackling this task should not pre-train on any data contained in the dataset. However, they may and are expected to make use of general pretraining on other data. In the case of our benchmark, for instance, pre-training on the Replica \cite{replica} or Matterport 3D datasets \cite{hm3d} is not allowed nor is using data containing the YCB~\cite{ycb} objects, as the objects are contained in our dataset.
\section{Dataset}
\label{sec:dataset}

\begin{figure*}[ht] 
  \centering
  \begin{minipage}[b]{0.2\textwidth}
    \centering
    \includegraphics[width=.99\textwidth]{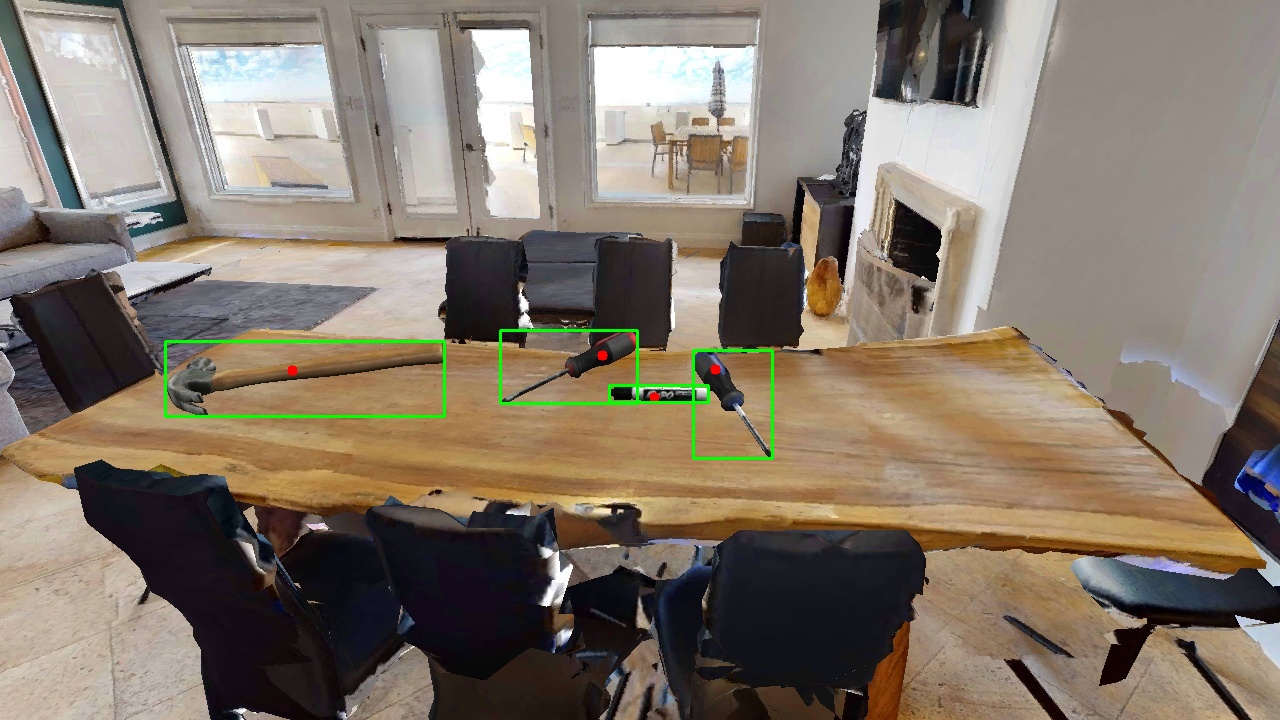} 
  \end{minipage}
  \begin{minipage}[b]{0.2\textwidth}
    \centering
    \includegraphics[width=.99\textwidth]{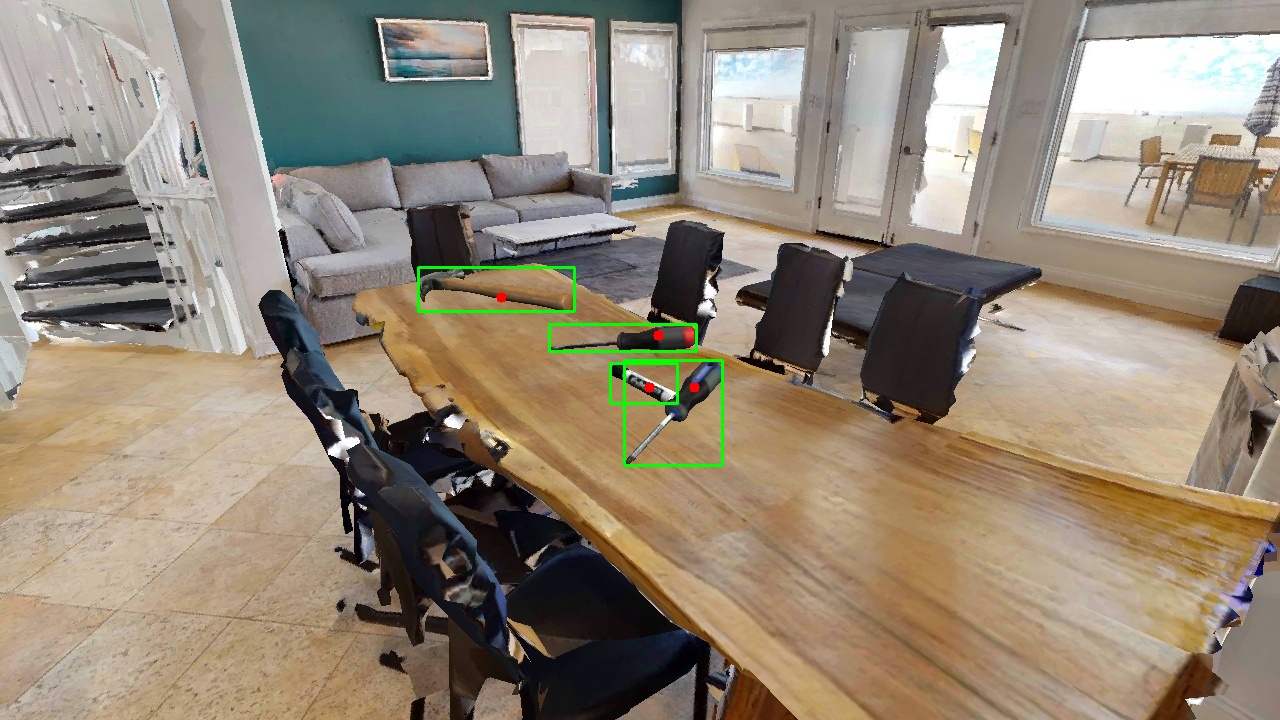}
  \end{minipage}
  \begin{minipage}[b]{0.2\textwidth}
    \centering
    \includegraphics[width=.99\textwidth]{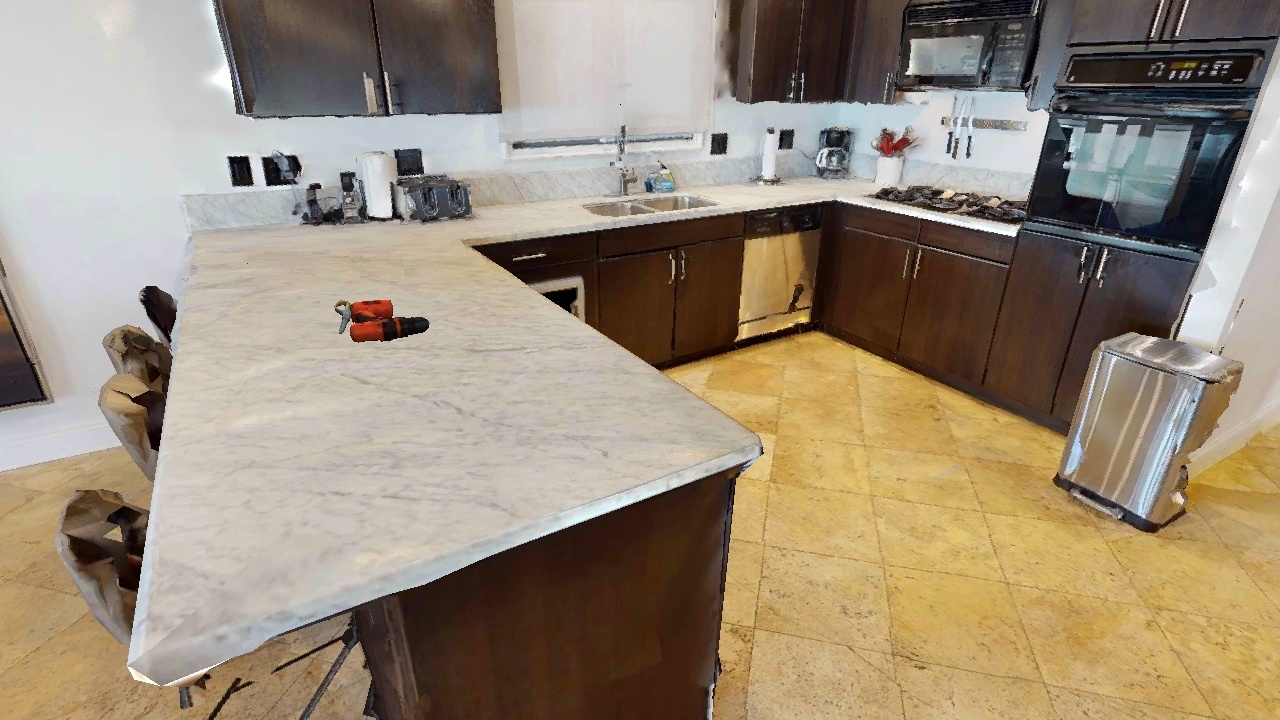}
  \end{minipage}
  \begin{minipage}[b]{0.2\textwidth}
    \centering
    \includegraphics[width=.99\textwidth]{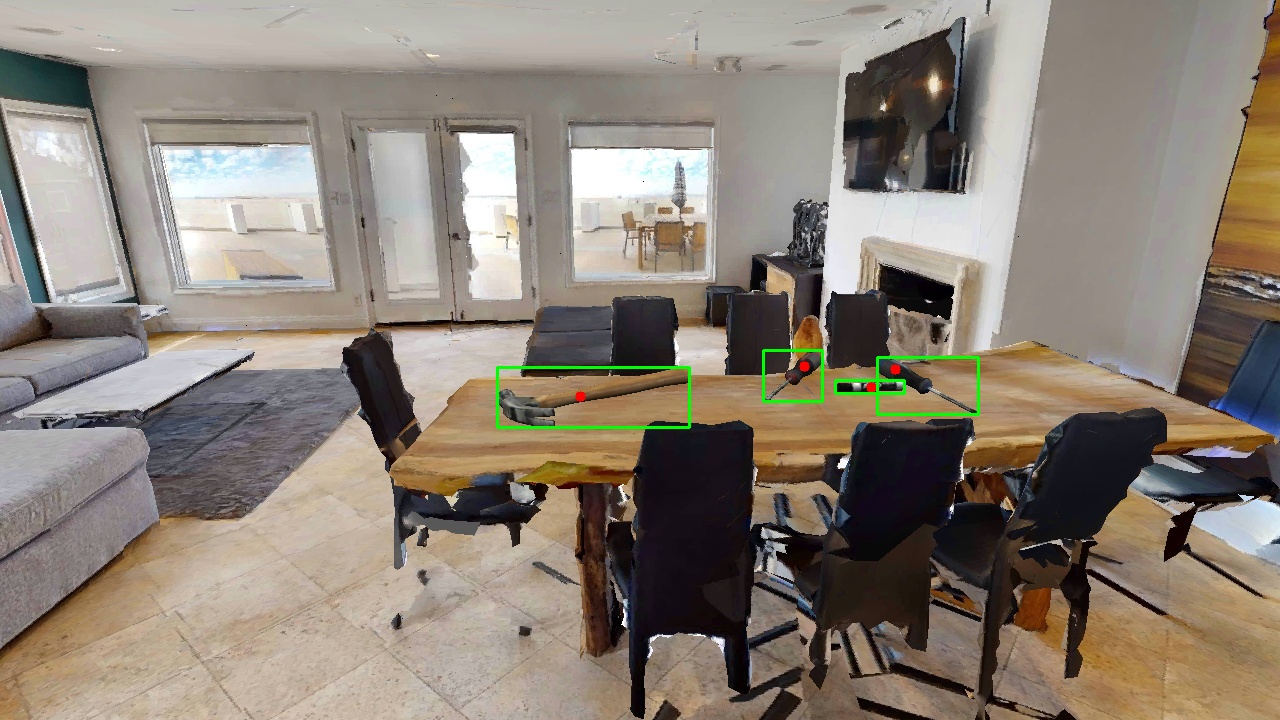}
  \end{minipage}
  \begin{minipage}[b]{0.2\textwidth}
    \centering
    \includegraphics[width=.99\textwidth]{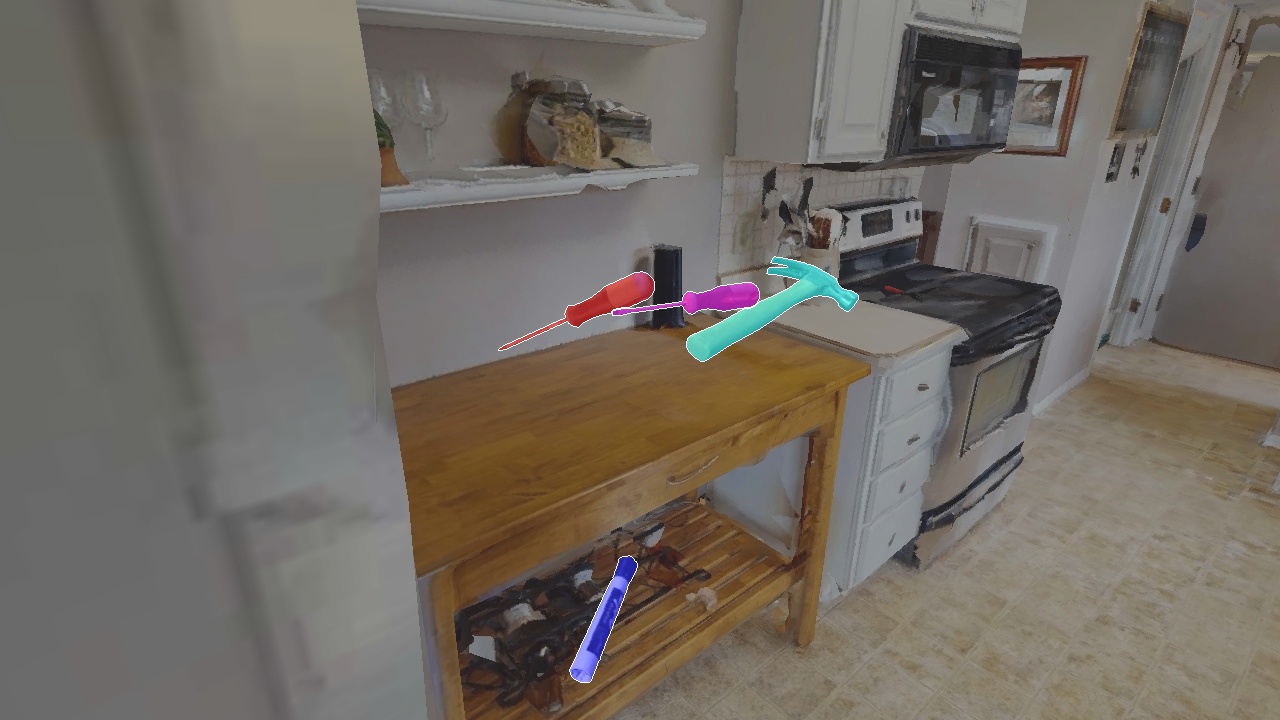}
  \end{minipage}
  \begin{minipage}[b]{0.2\textwidth}
    \centering
    \includegraphics[width=.99\textwidth]{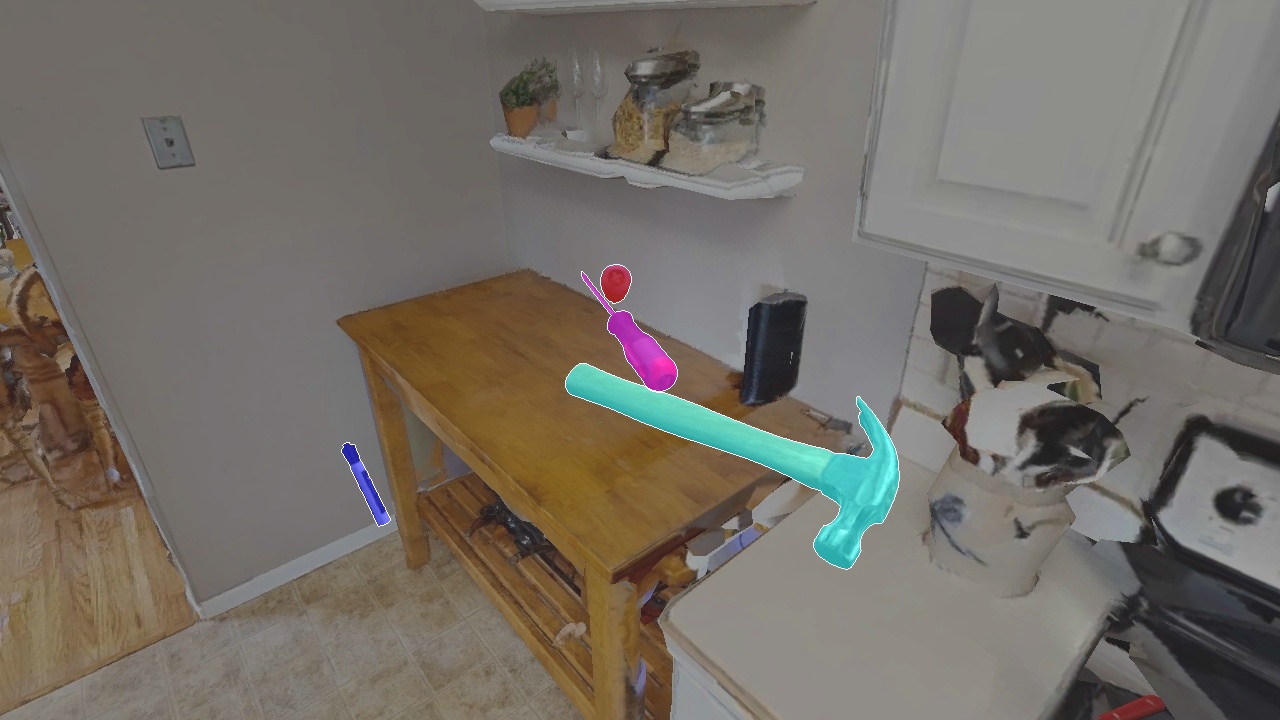}
  \end{minipage}
  \begin{minipage}[b]{0.2\textwidth}
    \centering
    \includegraphics[width=.99\textwidth]{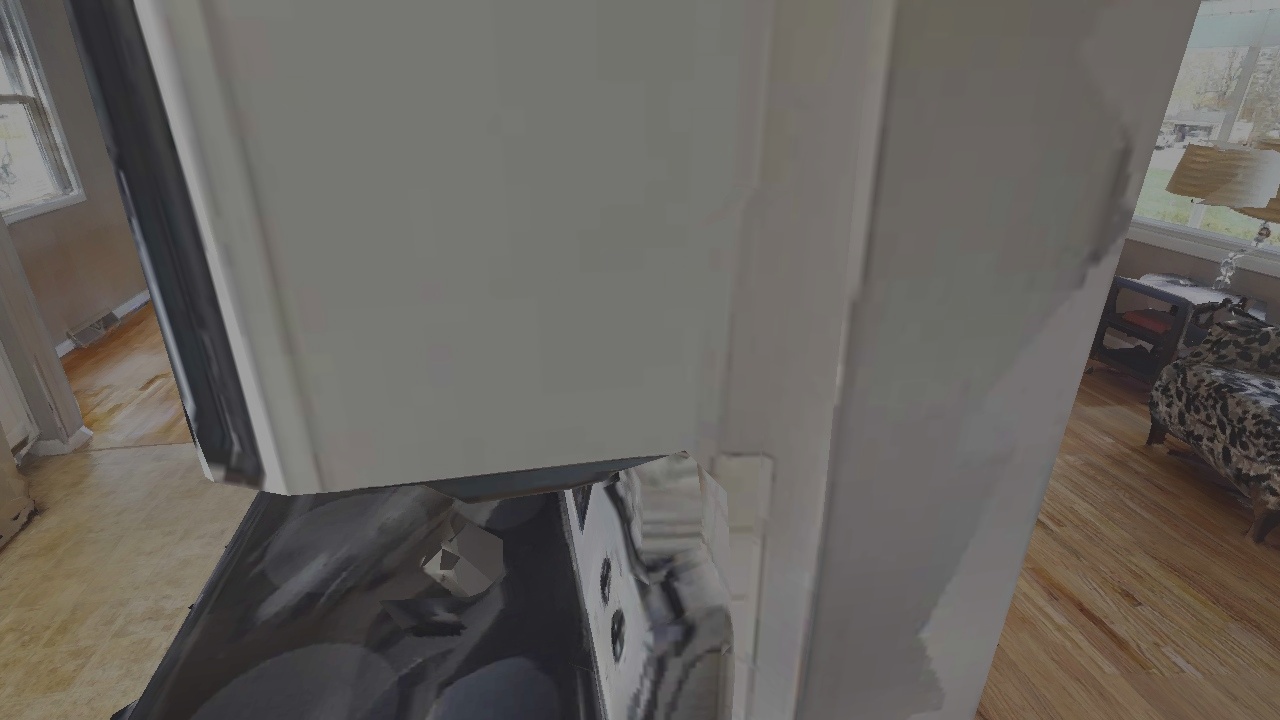}
  \end{minipage}
  \begin{minipage}[b]{0.2\textwidth}
    \centering
    \includegraphics[width=.99\textwidth]{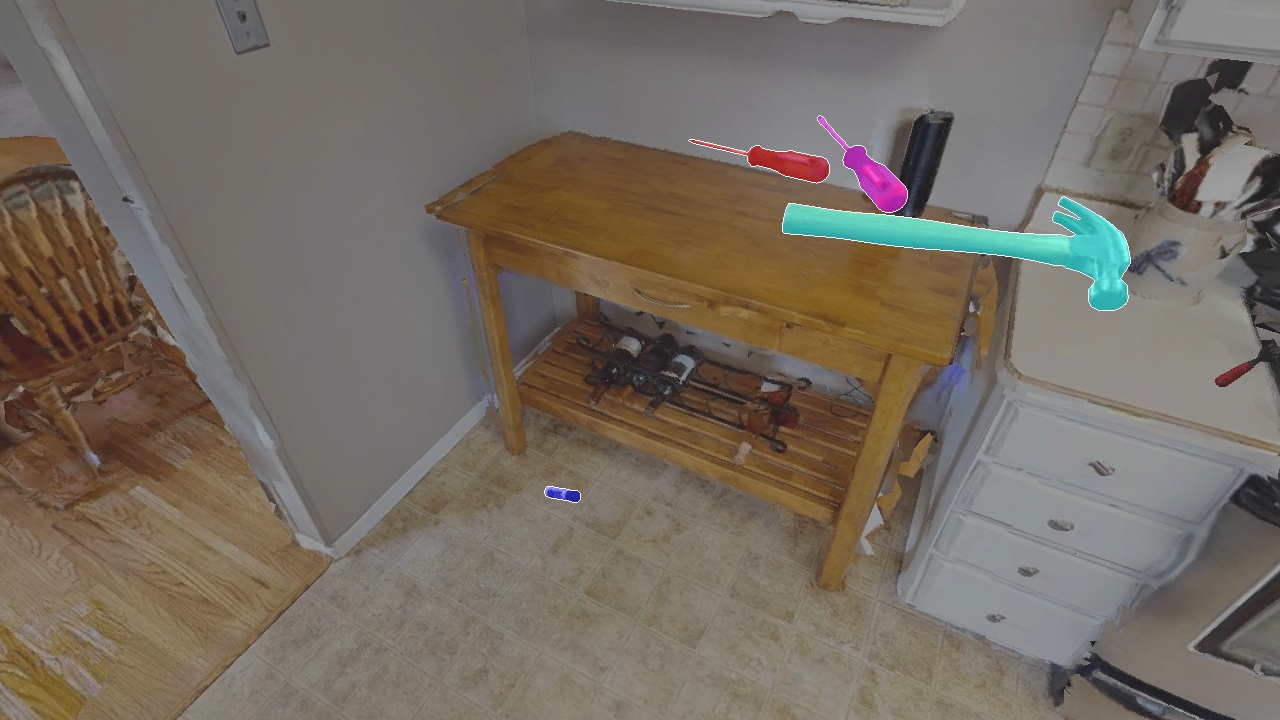}
  \end{minipage}
  \begin{minipage}[b]{0.2\textwidth}
    \centering
    \includegraphics[width=.99\textwidth]{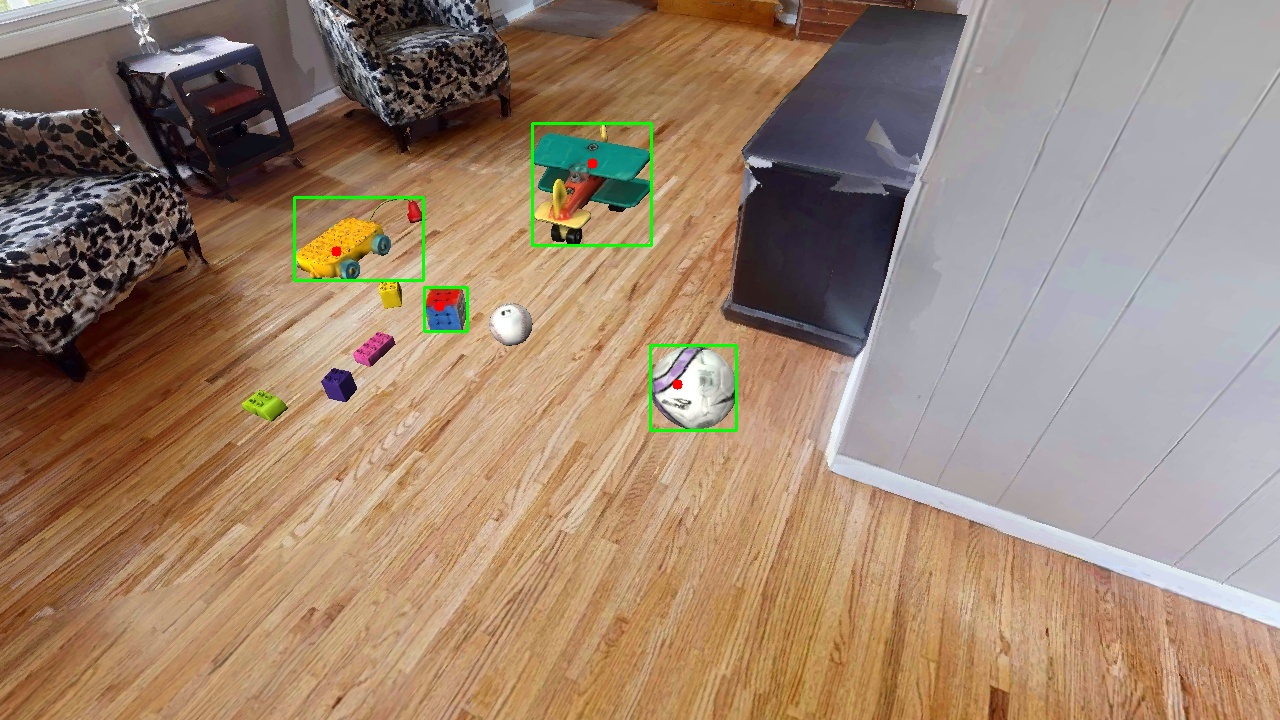}
  \end{minipage}
  \begin{minipage}[b]{0.2\textwidth}
    \centering
    \includegraphics[width=.99\textwidth]{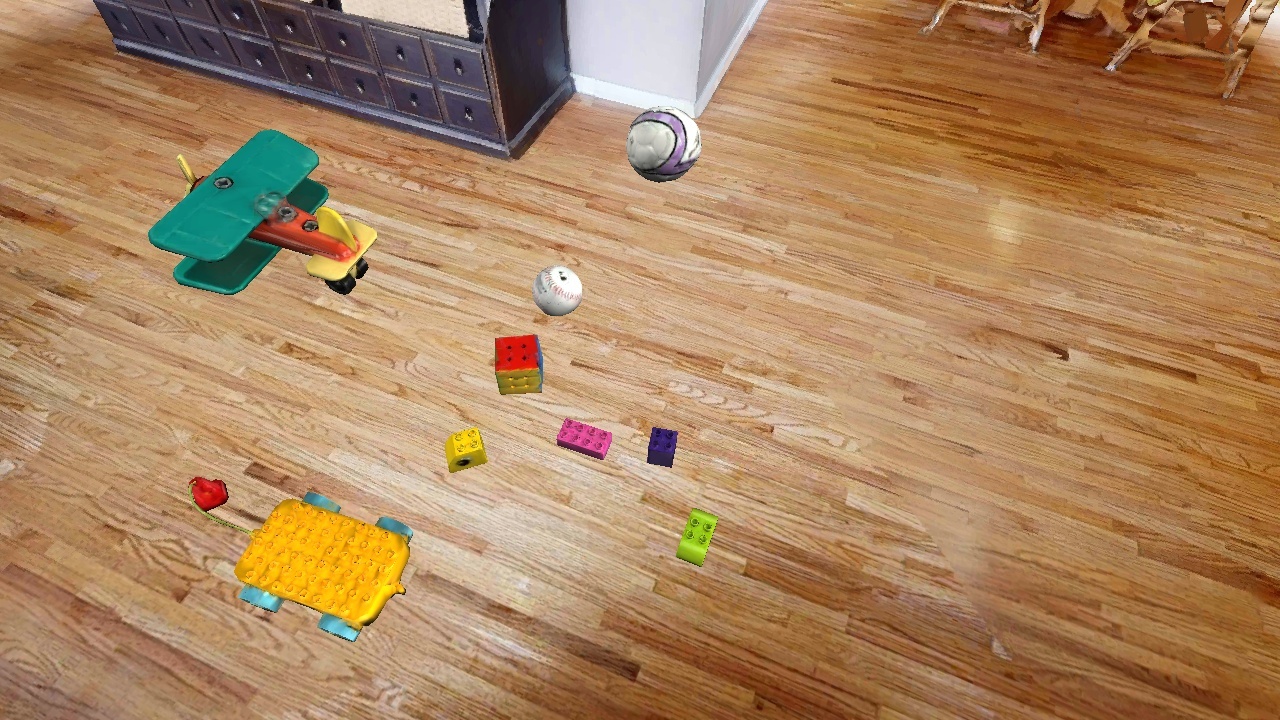}
  \end{minipage}
  \begin{minipage}[b]{0.2\textwidth}
    \centering
    \includegraphics[width=.99\textwidth]{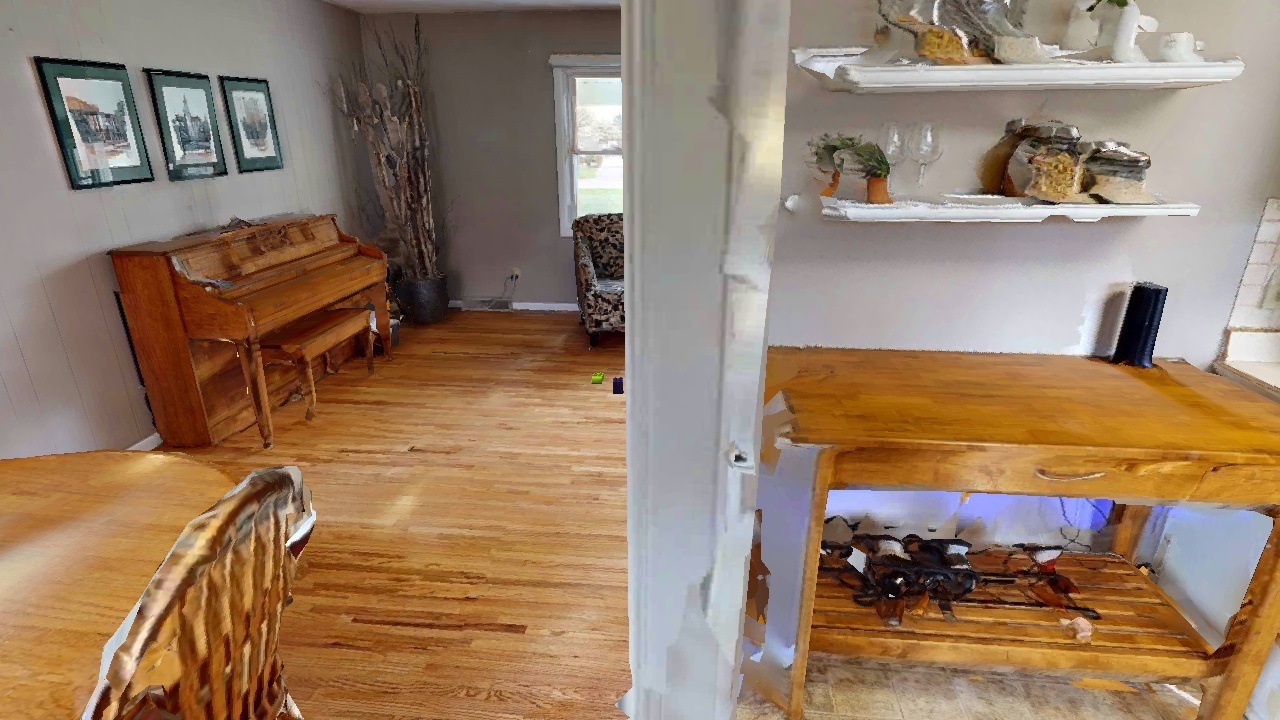}
  \end{minipage}
  \begin{minipage}[b]{0.2\textwidth}
    \centering
    \includegraphics[width=.99\textwidth]{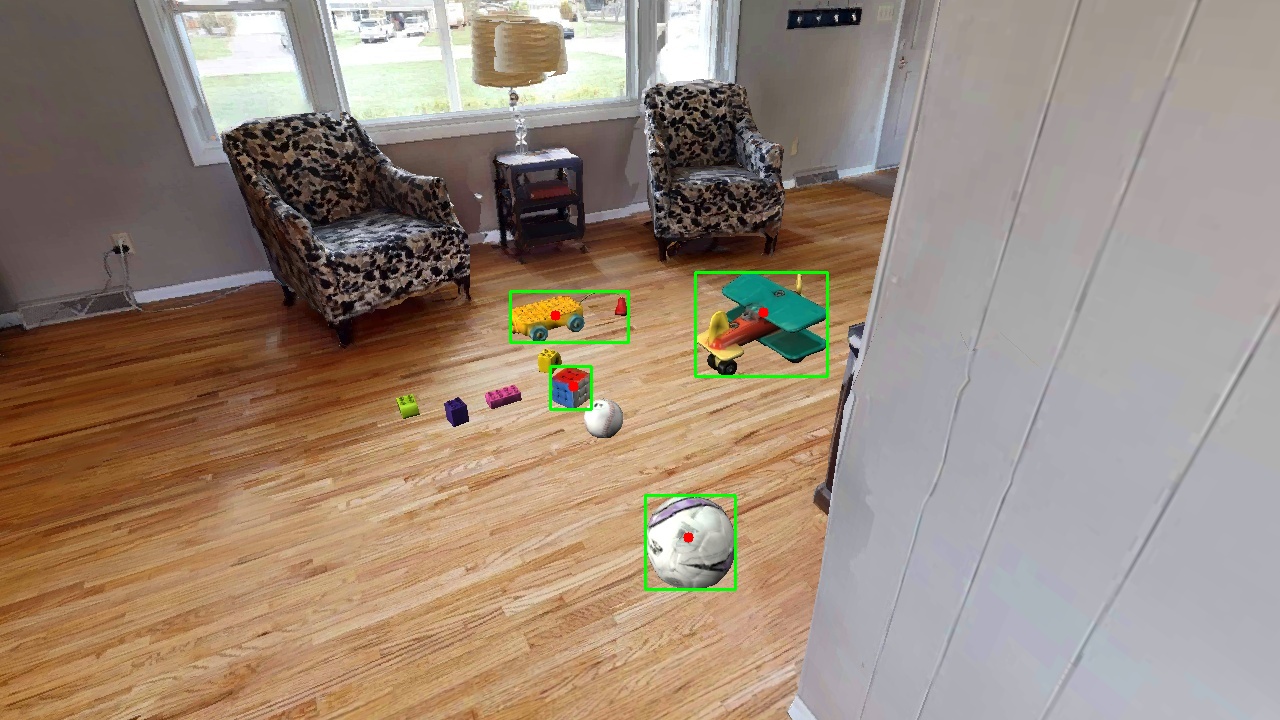}
  \end{minipage}
  \begin{minipage}[b]{0.2\textwidth}
    \centering
    \includegraphics[width=.99\textwidth]{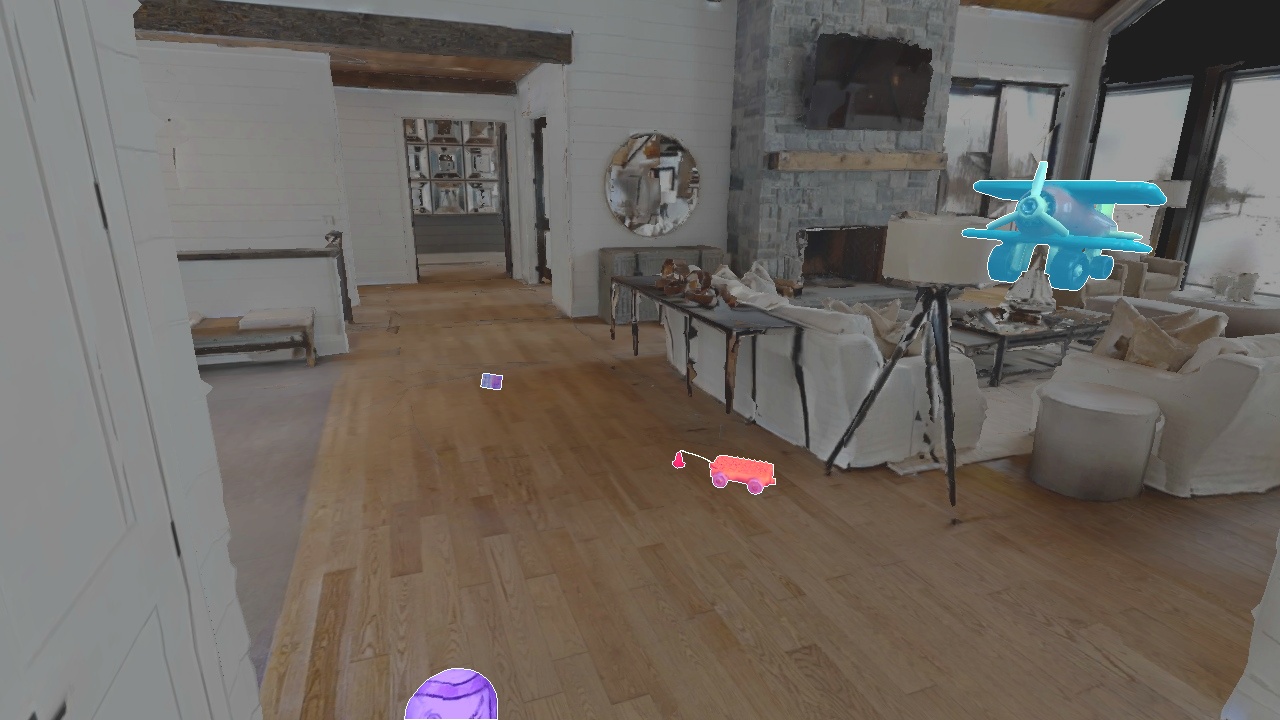}
  \end{minipage}
  \begin{minipage}[b]{0.2\textwidth}
    \centering
    \includegraphics[width=.99\textwidth]{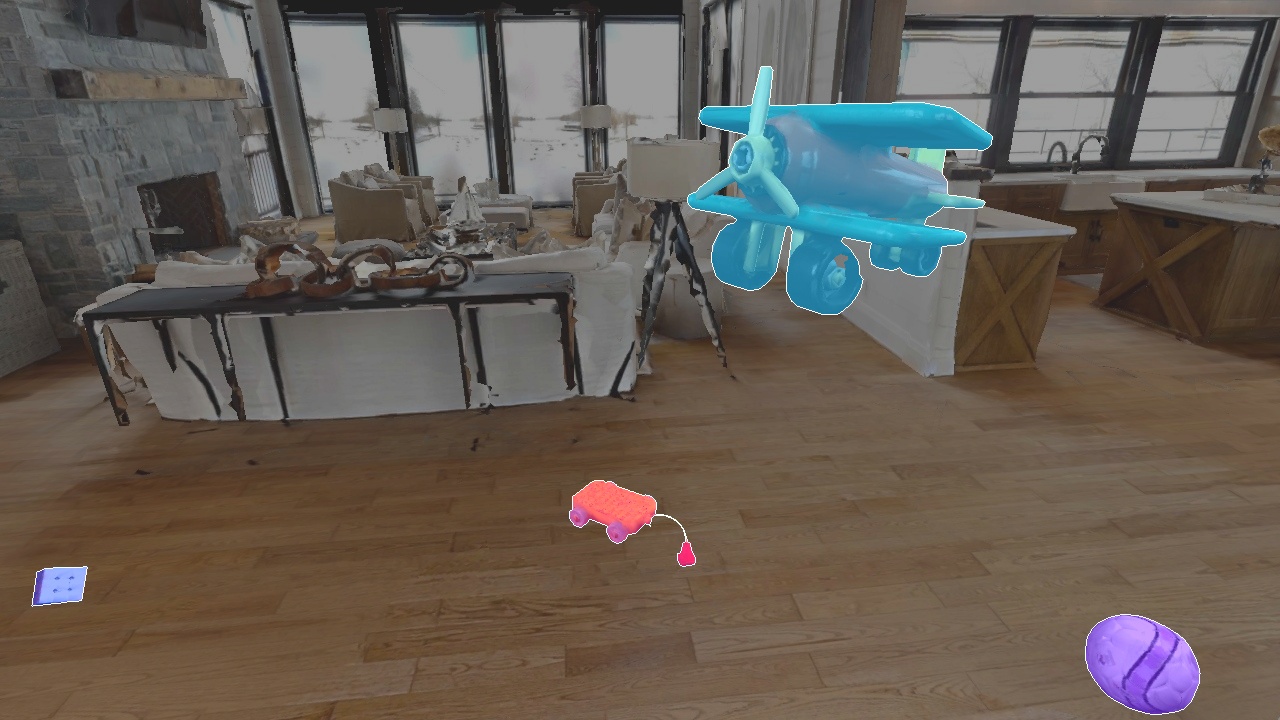}
  \end{minipage}
  \begin{minipage}[b]{0.2\textwidth}
    \centering
    \includegraphics[width=.99\textwidth]{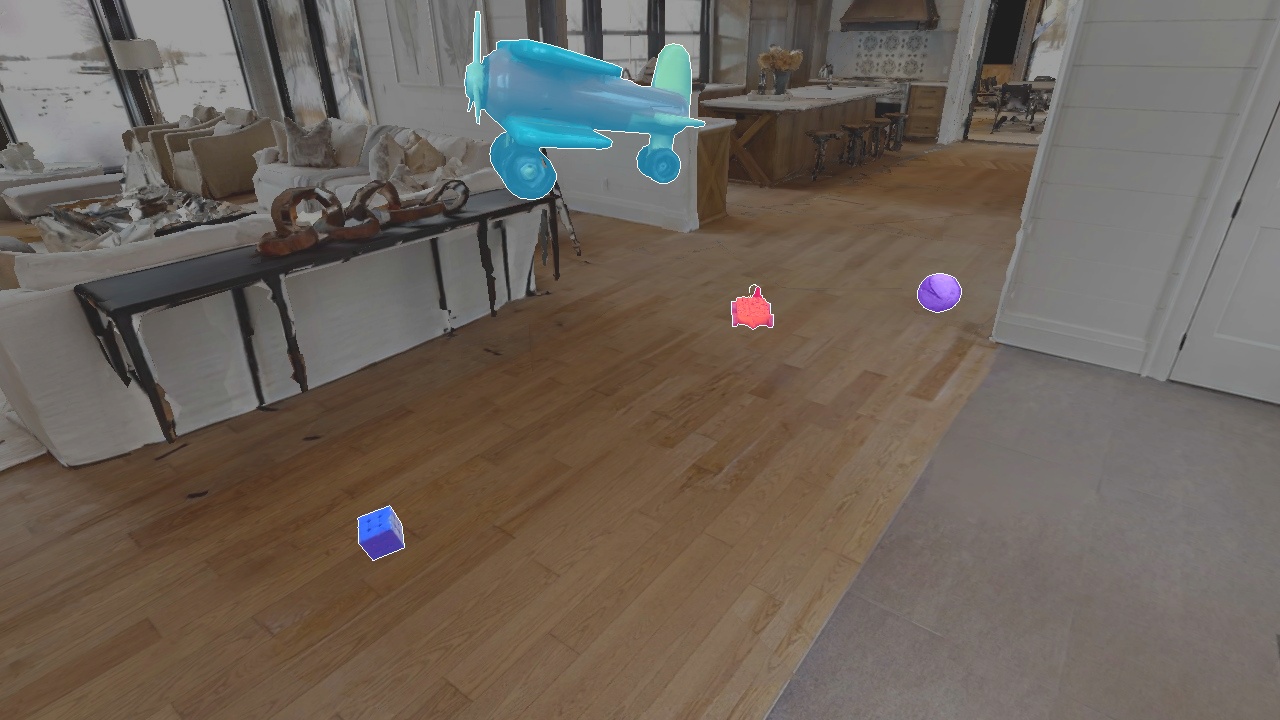}
  \end{minipage}
  \begin{minipage}[b]{0.2\textwidth}
    \centering
    \includegraphics[width=.99\textwidth]{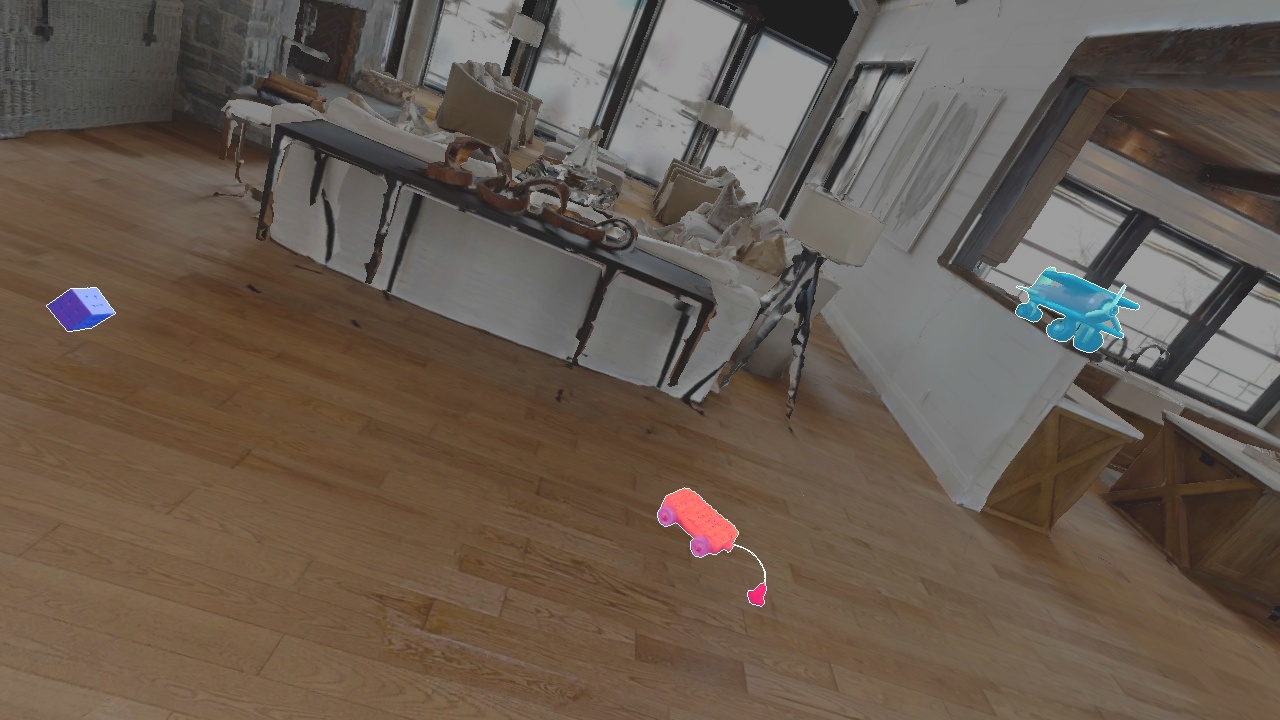}
  \end{minipage}
  \caption{Samples from the dataset. The first and third rows show examples of sparsely annotated train sequences. The second and fourth row show dense ground truth segmentations from the evaluation set. In both cases, the train sequence is static, while the shown evaluation sequence is dynamic (pen falling off the countertop; toy airplane flying through the room). Best viewed in color.}
  \label{fig:dataset_samples} 
\end{figure*}


\rev{The benchmark contains $24$ \emph{test cases}, each consisting of a training \rev{set} $\mathcal{D}^{\text{train}}$ of one sequence and an evaluation \rev{set} $\mathcal{D}^{\text{eval}}$ of $1$ to $5$ evaluation sequences. For each test case, annotations for a single-shot and a multi-shot scenario are provided. In the single-shot scenario, a single annotation per object is provided, while in the multi-shot scenario there are between $6$ and $14$ annotations per object, varying 
across test cases.
This is to test how different algorithms perform with a varying amount of supervision. Each test case contains up to $8$ objects to track. In total, the dataset consists of $84$ combined train and evaluation sequences. The length of each sequence is $400$ frames at a frame-rate of \SI{30}{\Hz}.
In~\Cref{tab:benchmark_size} we compare its size to commonly used VOS benchmark datasets. Note that, in contrast to these, our dataset is only meant for \emph{evaluating}, not training models and that,
unlike for instance YouTube-VOS~\cite{youtube_vos},
its focus
is not on achieving large scale.
In particular, the amount of data provided for each object is chosen so as to prevent using the training sequences to overfit to the characteristics of the benchmark.
}
\begin{table}[!htbp]
  \begin{center}
    {\small{
\begin{tabularx}{.9\linewidth}{l X s}
\toprule
Benchmark & \#~instances per sequence (avg.) & \# Annotations\\
\midrule
Ours & 3.52 & 124,681\\
DAVIS 2016\cite{davis} & 1 & 3,455\\
DAVIS 2017\cite{davis2017} & 2.56 & 10,474\\
YouTube-VOS\cite{youtube_vos} & 1.74 & 197,272\\
\bottomrule
\end{tabularx}
}}
\end{center}
\vspace{-5pt}
\caption{\rev{Comparison of dataset size to the three most common VOS benchmarks. \# Annotations is the total number of unique object annotations. \# instances per sequence (avg.) is the average number of annotated object instances per sequence}}
\label{tab:benchmark_size}
\end{table}

\rev{As our benchmark is designed to test single- and few-shot methods, we limit the amount of data available in training sequences and rather propose to test methods across a wide range of test scenarios. Methods tackling this task can make use of external data, as long as they do not contain data from the Matterport, Replica or YCB datasets.}

The scenes in which the video sequences are recorded are indoor scenes of the Replica~\cite{replica} and the Habitat-Matterport 3D research dataset~\cite{hm3d}. We use data from the YCB dataset~\cite{ycb} as objects to be segmented and re-identified. A few sample frames of the dataset can be seen in~\Cref{fig:dataset_samples}.

\subsection{Data Recording}
The scenes are recorded using the Habitat AI Simulator~\cite{habitat, habitat2}. To generate coherent sequences, we manually set the $6$-DoF poses of keyframes and subsequently interpolate the positions using B-splines and the orientation using spherical quadrangular interpolation of the quaternion orientations with cubic splines. This results in smooth trajectories with continuous linear and angular accelerations. Using semantic annotations of the Replica~\cite{replica} and Habitat Matterport 3D research dataset~\cite{hm3d} allows us to circumvent expensive manual labeling and yields pixel-perfect annotations while retaining near photo-realistic data.

\subsection{Attributes}

Taking inspiration from the DAVIS dataset~\cite{davis}, we assign attributes to every training/evaluation sequence. These attributes are chosen to point out the strengths and weaknesses of methods tackling the benchmark.
The attributes and their number of occurrences are shown in~\Cref{tab:scene_attributes}.

\begin{table}[!htbp]
  \begin{center}
    {\small{
\begin{tabularx}{.9\linewidth}{l X r}
\toprule
ID & Description & \#\\
\midrule
DYN & The scene contains dynamic objects & $26$\\
CLT & The scene is cluttered with other objects & $78$\\
CLA & The scene contains multiple distinct objects of the same class (\eg, 2 different cans) & $43$\\
SML & The mean ratio between object bounding box and image area is smaller than $0.005$ & $58$\\
SMF & There are frames with a ratio between the object bounding box and the image area smaller than $0.001$ & $73$\\
FST & The recording contains fast camera movements\rev{, defined as linear velocity $>\SI{1.5}{\meter/\second}$ or angular velocity $>\SI{1.0}{\radian/\second}$}& $64$\\
\bottomrule
\end{tabularx}
}}
\vspace{-5pt}
\end{center}
\caption{Attributes of the dataset scenes with associated descriptions and number of occurrences.}
\label{tab:scene_attributes}
\end{table}

\section{Evaluation}
\label{sec:evaluation}

To facilitate the comparison of different methods, we first draw a distinction between two possible scenarios: one where the object is visible within a frame, and the other where the object is not present in the frame. This categorization is designed so that we can separately compute evaluation metrics for each scenario.
In the scenario where the object instance $k$ is visible (the number of pixels in the mask $\mathbf{M}_{k,j,gt}(\hat{t})$ is larger than zero), we compute four different metrics: 
\begin{enumerate}
    \item The \emph{Jaccard index} $\mathcal{J}$(Intersection over Union) defined by 
        \begin{equation}
            \mathcal{J} = \frac{\mathbf{M}_{k,j}(\hat{t}) \cap \mathbf{M}_{k,j,gt}(\hat{t})}{\mathbf{M}_{k,j}(\hat{t}) \cup \mathbf{M}_{k,j,gt}(\hat{t})} \in [0,1]
        \end{equation}
measures the accuracy of the mask prediction, when the object is visible. Here, $\mathbf{M}_{k,j}(\hat{t})$ is the predicted binary segmentation mask of the object in a single frame and $\mathbf{M}_{k,j,gt}(\hat{t})$ is the corresponding ground truth mask.
\item The \emph{boundary $F_1$-score} $\mathcal{F}$. For this we define the boundary of a segmentation mask $\mathbf{B}$ as $\boldsymbol{\partial}\mathbf{B} = \overline{\mathbf{B}} \setminus \mathbf{B}^\mathrm{o}$\rev{, where $\overline{\mathbf{B}}$ denotes the closure of $\mathbf{B}$ and $\mathbf{B}^\mathrm{o}$ its interior}. These are the boundary pixels of the segmentation masks. The $F_1$-score of the boundary of the two segmentation masks is calculated with
\begin{equation}
    \mathcal{F} = \frac{2~\text{TP}}{2~\text{TP} + \text{FP} + \text{FN}},
\end{equation}
where $\text{TP} = |\boldsymbol{\partial} \mathbf{M}_{k,j}(\hat{t}) \cap \boldsymbol{\partial} \mathbf{M}_{k,j,gt}(\hat{t})|$, $\text{FP} = |\{\boldsymbol{\partial} \mathbf{M}_{k,j}(\hat{t}) \notin \boldsymbol{\partial} \mathbf{M}_{k,j,gt}(\hat{t})\}|$ and $\text{FN} = |\{\boldsymbol{\partial} \mathbf{M}_{k,j,gt}(\hat{t}) \notin \boldsymbol{\partial} \mathbf{M}_{k,j}(\hat{t})\}|$. In the scenario of spatial perception, this measure is of particular interest, as it quantifies to what degree a mask either leaks into the background or undershoots and fails to cover the entire object. Preventing such leakage is crucial for building up a good 3D representation of an object.
\end{enumerate}
Both for $\mathcal{J}$ as well as $\mathcal{F}$ we calculate the mean over an entire sequence and report this as the final measure, with higher values indicating better performance. These two measures follow DAVIS~\cite{davis}. Additionally, we define:
\begin{enumerate}
  \setcounter{enumi}{2}
  \item The \emph{visible misclassification rate} $J_{\text{mis, v}}$ as the ratio of frames, where $\mathcal{J} < t_\text{mis} = 0.4$. This measures the ratio of frames where at most a small part of the object was correctly classified.
  \item The \emph{visible false detection rate} $J_{\text{fd, v}}$: the ratio of frames, where $\mathcal{J} < t_\text{fd} = 0.1$ and $|\mathbf{M}_{k,j}| > |\mathbf{M}_{k,j,gt}| \times \mathcal{J}$. This measures the ratio of frames in which an object that is not the same instance as the correct object is segmented, even though the correct object is visible.
\end{enumerate}
When the object is not visible ($|\mathbf{M}_{k,j,gt}| = 0$), we measure the \emph{non-visible false detection rate} $J_{\text{fd, n}}$ as the ratio of frames, where $|\mathbf{M}_{k,j}| > 0$.

\section{Baseline Method}

We provide a baseline method for the benchmark that methods can be compared against. It relies on building feature descriptors of an object instance and on instance segments from the Segment Anything Model (SAM)~\cite{sam}. The feature descriptors are used to re-identify the objects in other scene contexts by constructing a simple classifier for each object.

In this section we describe our method, provide an evaluation on our benchmark and perform an ablation study of the different components of our method.

\subsection{Pipeline}
\label{sec:method_pipeline}

\begin{figure}[!ht]
    \centering
    \includegraphics[width=.9\linewidth]{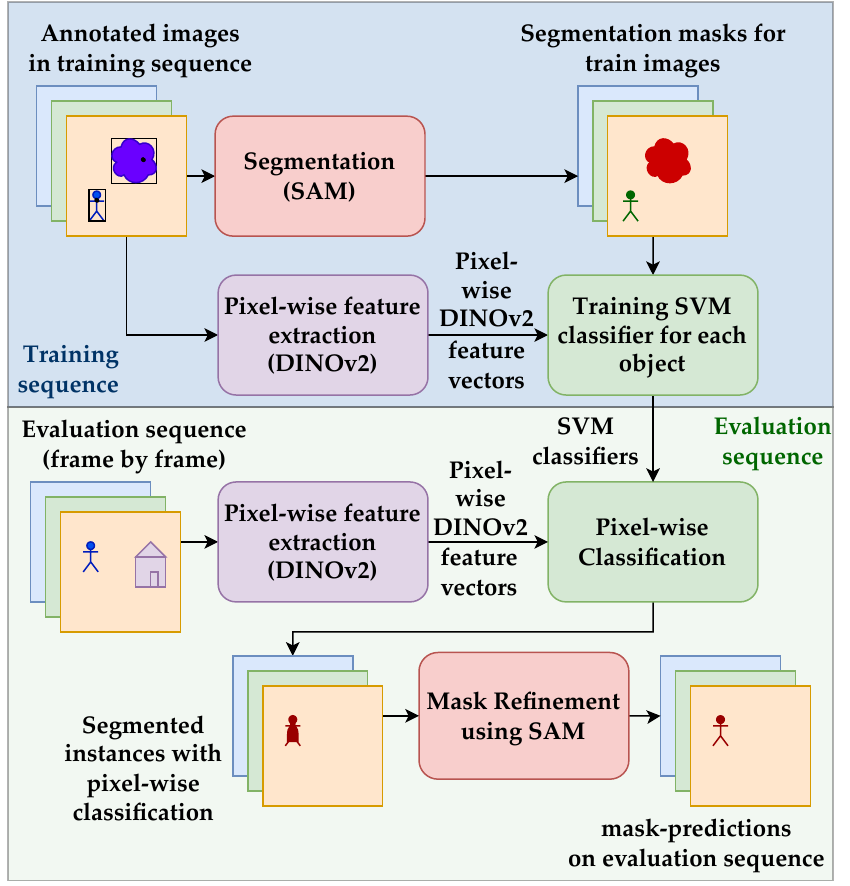}
    \caption{Pipeline of the baseline method. First, on annotated images of a training sequence, the point and bounding box prompts are transformed into masks using SAM~\cite{sam}. Second, DINOv2~\cite{dinov2} features are extracted on these images. Using the masks, a dataset of positive and negative DINOv2 features is created for each annotated object. This dataset is used to train a linear SVM for each object. These linear SVMs are sequentially applied to the dense upsampled DINOv2 features of images in evaluation sequences to determine instance membership. Finally, the resulting masks are refined with SAM.}
    \label{fig:baseline_method_pipeline}
    \vspace{-15pt}
\end{figure}

Our method builds on pixel-wise features, which we use to describe our objects. An overview of the pipeline can be seen in ~\Cref{fig:baseline_method_pipeline}. 

In the initial step, SAM is prompted with the bounding box and point prompt of the annotations in the training sequence. Inspired by~\cite{baking_in_the_feature}, we leverage image features learned on a massive, diverse dataset to describe object instances. Therefore, each of the images is passed through a pretrained DINOv2~\cite{dinov2} backbone to extract a feature tensor $T_\textrm{DINO} \in \mathbb{R}^{40  \times40\times1024}$. This tensor is upsampled to the image size, using bicubic upsampling. A linear maximal margin classifier (SVM) is trained for each object, using the dense DINOv2 features that lie within the predicted mask on the train images as positive samples; as negative samples, we use DINOv2 feature vectors sampled from outside of the predicted mask. \rev{To prevent overfitting to the current scene context, we include a set of DINOv2 feature vectors previously extracted from a diverse set of random images as negative examples.}
For frames in the evaluation dataset, we similarly compute upsampled DINOv2 feature vectors and use the classifier to determine instance membership pixel-wise. Optionally, the resulting masks are refined using SAM. \rev{To do this, we compute the dot product between all (pixel-wise) features and the SVM weights for an object. This effectively scores each pixel in terms of similarity to the object instance. Then, we select the $5$ largest local maxima of this dot product to prompt SAM. This results in multiple mask proposals by SAM.} These mask proposals are allocated to the object instances using linear assignment, maximizing \rev{the dot product of the SVM weights for an object and} the feature vectors that lie beneath a mask proposal.
Finally, if the average of the features that are covered by a mask lie on the negative side of the decision boundary of an assigned SVM, the mask is discarded.

The benefits of the method include the fact that it does not rely on external training with segmentation or video data and purely uses out of the box models. Further, it is agnostic to the class of the objects.


\subsection{Quantitative Results}
\begin{figure}[!ht]
\begin{center}
   \includegraphics[width=.9\linewidth]{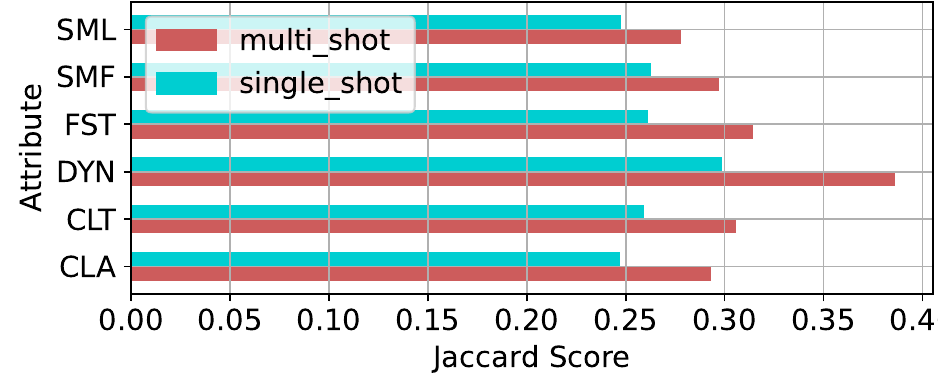}
\end{center}
    \caption{\rev{Mean $\mathcal{J}$ score of the baseline method with respect to scene attributes.}}
\label{fig:dataset_mious}
\end{figure}

We evaluate the baseline method on the proposed benchmark. \rev{The $\mathcal{J}$ with respect to scene attributes can be seen in~\Cref{fig:dataset_mious}}. The method performs best in scenarios with little clutter and no objects of the same semantic class and achieves lowest performance in scenes with \rev{clutter, objects of the same class, and small objects}.
The evaluation metrics averaged over all evaluation sequences can be seen in~\Cref{tab:method_scores}.

\begin{table}[!htbp]
  \begin{center}
    {\small{
\begin{tabularx}{0.9\linewidth}{l X X X X X}
\toprule
 & $\mathcal{J}$ & $\mathcal{F}$ & $J_{fd,v}$ & $J_{fd,n}$ & $J_{mis}$ \\
\midrule
single shot & $0.2686$ & $0.1249$ & $0.1319$ & $0.0998$ & $0.7092$\\
multi shot & $0.3224$ & $0.1474$ & $0.0629$ & $0.0401$ & $0.6545$\\
\bottomrule
\end{tabularx}
}}
\end{center}
\caption{Performance of the baseline method expressed in the metrics of~\Cref{sec:evaluation}. \rev{Averaged over all evaluation sequences.}}
\label{tab:method_scores}
\end{table}

\subsection{Qualitative Results}

\begin{figure}[ht] 
\centering
  \begin{minipage}[b]{0.45\linewidth}
    \centering
    \includegraphics[width=.99\textwidth]{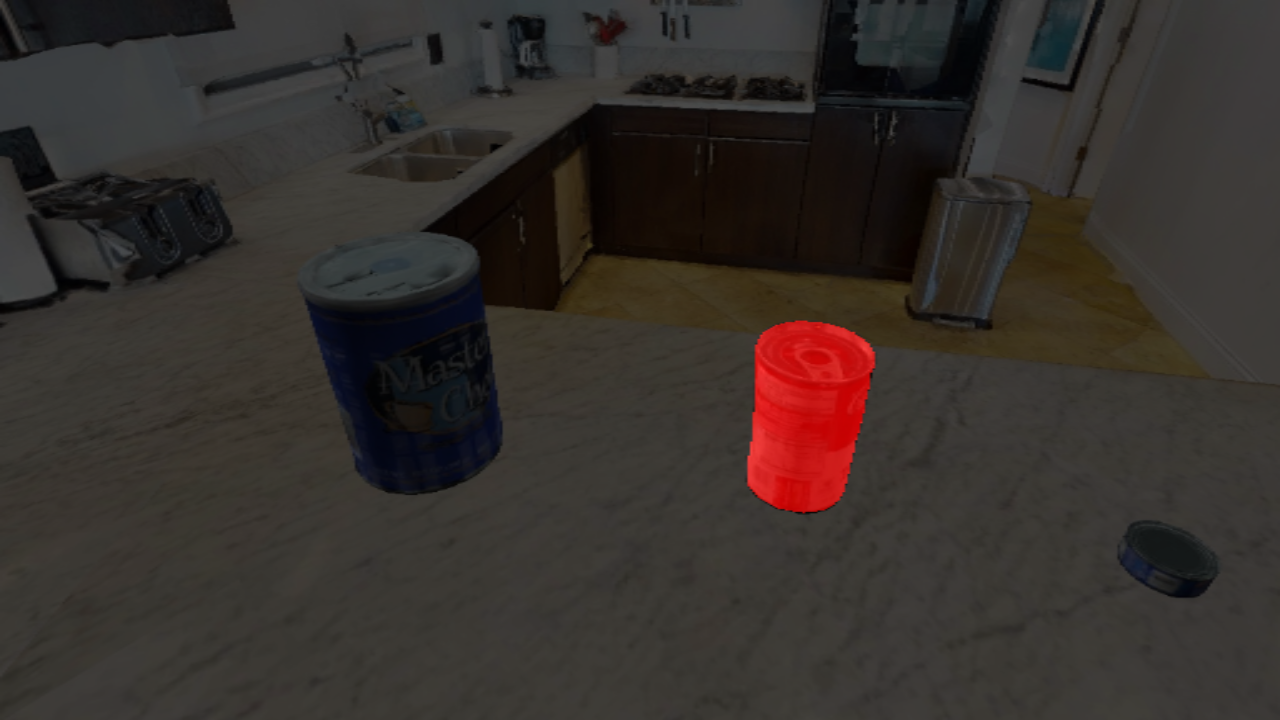} 
  \end{minipage}
  \begin{minipage}[b]{0.45\linewidth}
    \centering
    \includegraphics[width=.99\textwidth]{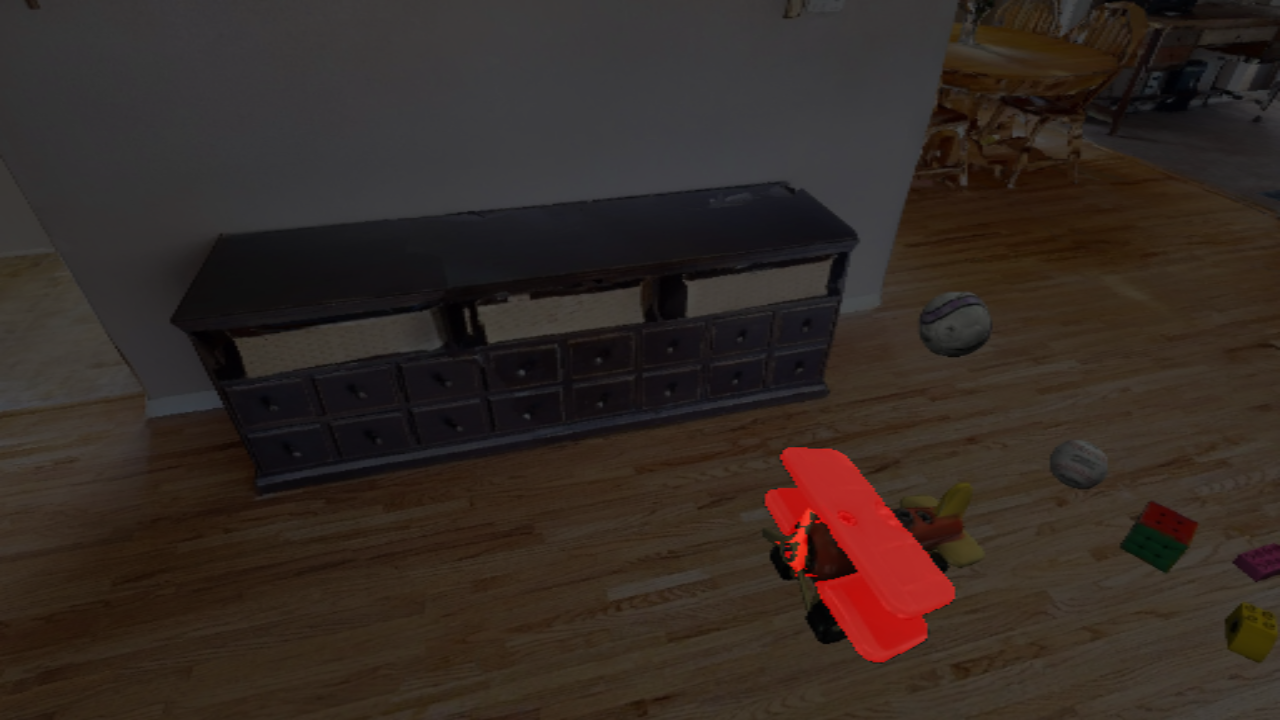} 
  \end{minipage} 
  \centering
  \caption{Qualitative results showing the effectiveness of SAM as a way to transform the given prompts into initial segmentations. Left: good example, right: bad example.}
\label{fig:qualitative_results_initial_mask_prediction} 
\end{figure}

Using SAM to transform the point and bounding-box prompts into masks in general works well, however even small failures lead to inconsistencies over the entire evaluation sequence, as no additional data from the training sequence is used. A good and a bad example can be seen in~\Cref{fig:qualitative_results_initial_mask_prediction}. This is especially apparent when the object is composed of multiple parts. Performance at this stage is critical, as negative features to train the SVM are sampled from the area outside the mask. Therefore, a mask that does not capture the entire object might lead to failure later on. However, this was observed in few sequences.

While the segmentation quality on the evaluation data is high (in frames where the correct instance is segmented, the $\mathcal{J}$ score is generally above $0.9$), the mask quality varies from frame to frame, as no temporal information is taken advantage of.

\subsection{Failure Cases}

The high rate of misclassifications (frames where $\mathcal{J} < 0.4$) shows that the robustness of the method can be improved. Especially in scenarios where the object is small and few pixels are available, the DINOv2-model~\cite{dinov2} fails to extract enough meaningful information to re-identify an object. Further, as the segmentation method (SAM~\cite{sam}) is not optimized with respect to the the boundary $F_1$-measure $\mathcal{F}$, the latter is very low. 

Moreover, as addressed in the previous paragraph, predictions in two subsequent frames may be entirely different due to no temporal knowledge being used.

Lastly, in the single-shot setting the baseline sometimes completely fails with objects of the same semantic class (for instance it fails in a \rev{test case} that requires distinguishing between a flat and a crosshead screwdriver that are both small in the image).
This is caused by the SVM classifier not being aware of the other instance in the scene. Therefore, the features of the very similar looking screwdriver might lie on the same side of the decision boundary as the one to track, as no negative examples of the same class exist.
A way to improve this would be to combine the method with state-of-the-art VOS methods (\eg,~\cite{xmem, qdmn, wang_look_2022}) to track an object while it is visible in subsequent frames. This then may be used in the training sequence to generate additional data for the object's SVMs to be trained on. It could also be beneficial in the evaluation sequences to improve robustness.

\subsection{Ablation on Method Components}

We perform ablations on the different components in our baseline method. First, we discuss the choice of object descriptors. We test CLIP~\cite{clip}, SAM~\cite{sam}, and DINOv2~\cite{dinov2} features. For the SAM features, we used a similar pipeline as in~\Cref{sec:method_pipeline}, simply replacing the DINOv2 feature tensor $T_\textrm{DINO}$ with the result of the SAM backbone $T_\textrm{SAM} \in \mathbb{R}^{64\times64\times256}$.

To be able to use CLIP feature vectors as descriptors, we take a different approach, as they are global and not local like the features resulting from the SAM or DINOv2 backbones.
This method relies on CLIP features~\cite{clip} extracted from object bounding box proposals. The bounding box proposals are generated by OW-DETR~\cite{ow_detr} which is a method for open world object detection based on DETR~\cite{detr}.
First, on annotated frames of the training sequence, CLIP features are extracted from the image cropped at the bounding box annotation. For each frame in the evaluation sequence OW-DETR bounding box proposals are generated. CLIP features are extracted from each bounding box and compared under the cosine similarity metric with the train features. The bounding box corresponding to the features that achieve the highest similarity is selected. If the maximum similarity $S_{c,max} \in \mathbb{R}$ is smaller than a threshold $t_{sim} \in \mathbb{R}$, no object is selected. From the selected bounding box, a segmentation mask is inferred by prompting SAM with the bounding box. 

We noticed that this performs badly when object instances of the same or a similar semantic class are present, as CLIP features mostly encode class-level knowledge. Further, it is sensitive to viewpoint changes, struggles with small objects and the best threshold $t_{sim}$ varies by the type of object.

\begin{figure}[ht] 
\centering
\begin{minipage}[c]{0.30\linewidth}
  \centering
  \includegraphics[width=.99\textwidth]{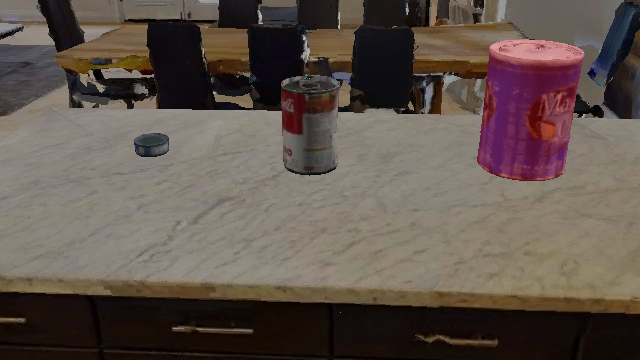}
\end{minipage}
\begin{minipage}[c]{0.30\linewidth}
  \centering
  \includegraphics[width=.99\textwidth]{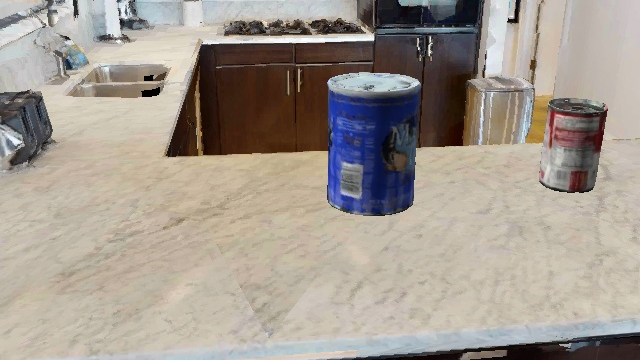}
\end{minipage}
\begin{minipage}[c]{0.30\linewidth}
  \centering
  \includegraphics[width=.99\textwidth]{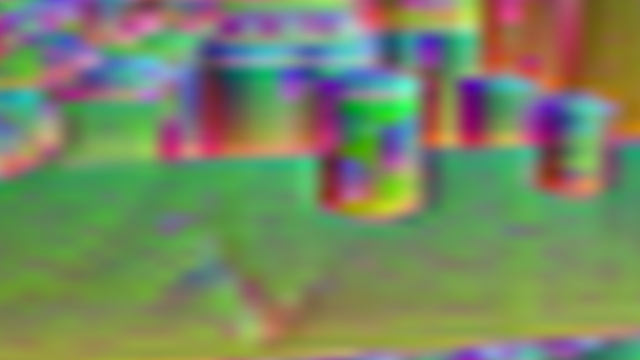}
  \includegraphics[width=.99\textwidth]{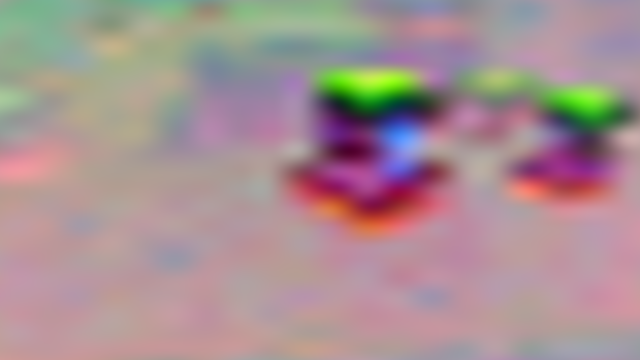}
\end{minipage}

\caption{An example of how different models encode instance-level knowledge. Left: Image with segmentation mask used to fit a PCA transformation. Middle: Inference image (different perspective). top right: Largest three principal components of the upsampled SAM features, visualized on the entire image. bottom right: Largest three principal components of the upsampled DINOv2 features, visualized on entire image. Note that the DINOv2 features better encode the class- and instance membership of the pixels of the larger can.}
\label{fig:pca_example}
\end{figure}

In contrast to the SAM and CLIP features, DINOv2 features used in combination with a SVM as described above are a better fit not only for distinguishing objects belonging to two different semantic classes, but also for differentiating between two objects of the same class. This approach using DINOv2 features consistently outperforms SAM and CLIP features on almost every sequence in the benchmark.
The discrepancy in how these features encode semantic knowledge can be shown by visualizing the first three principal components of the pixel-wise upsampled feature vectors in an RGB image. The dimensionality reduction is fit to the masked features of an object instance. An example can be seen in ~\Cref{fig:pca_example}. The DINOv2 features used in the way that we outlined seem to be significantly better object descriptors than the SAM features. 

\begin{figure}[ht] 
\centering
  \begin{minipage}[b]{0.45\linewidth}
    \centering
    \includegraphics[width=.99\textwidth]{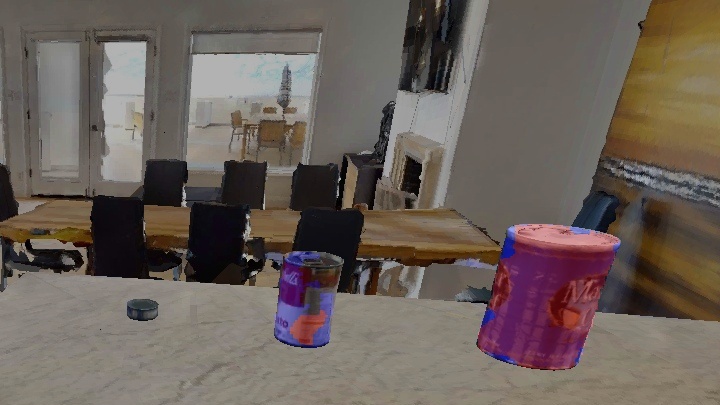} 
  \end{minipage}
  \begin{minipage}[b]{0.45\linewidth}
    \centering
    \includegraphics[width=.99\textwidth]{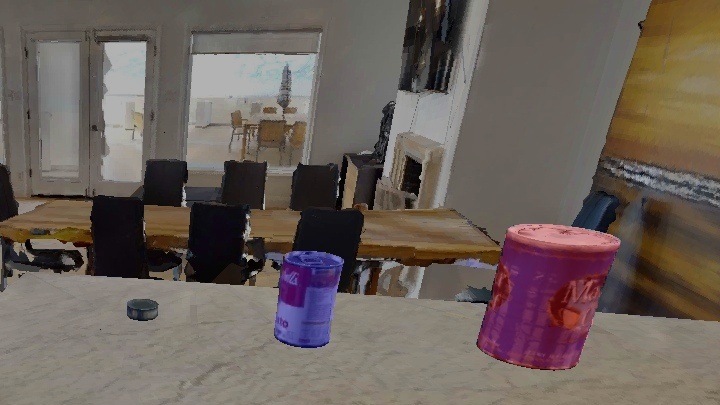} 
  \end{minipage} 
  \centering
  \begin{minipage}[b]{0.45\linewidth}
    \centering
    \includegraphics[width=.99\textwidth]{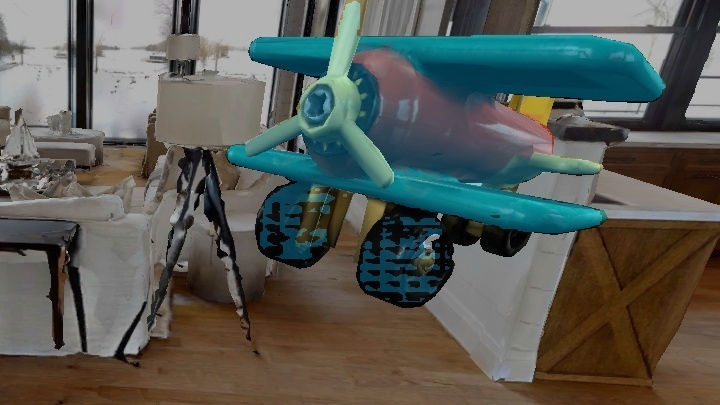} 
  \end{minipage}
  \begin{minipage}[b]{0.45\linewidth}
    \centering
    \includegraphics[width=.99\textwidth]{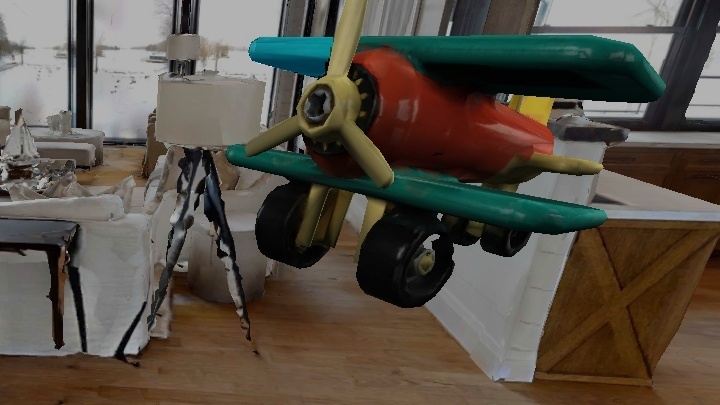} 
  \end{minipage}
  \caption{Selected examples of the benefits and drawbacks of using the final mask-refinement step with SAM. Left: Mask predictions using no SAM refinement. Right: Mask predictions with SAM refinement. Notice that SAM refinement helps with multiple object instances of the same class, but decreases the performance when the object consists of multiple parts.}
  \label{fig:sam_refinement} 
\end{figure}

Additionally, we discuss the use of the mask-refinement step using SAM outlined in \Cref{sec:method_pipeline}. Performing this step greatly improves the boundary $F_1$-score $\mathcal{F}$ and increases the method's ability to deal with objects of the same class. However, it struggles with objects that are assembled of multiple parts, as SAM proposes masks for each part individually. An example illustrating the strengths and weaknesses of the mask-refinement step can be seen in \Cref{fig:sam_refinement}.

\section{Conclusion}
We presented a benchmark for few-shot Video Object Instance Segmentation and Re-Identification.
The goal of the benchmark is to improve object segmentation and re-identification in different scene contexts. It offers various different scenarios: single-object/multi-object segmentation, single-shot/few-shot annotation data in the training sequences, RGB/RGB-D data and $6$-DoF camera poses/no camera poses available to the user. The dataset and evaluation code will be released upon publication.

We also developed a method as a baseline solution for the task, tackling the problem by forming DINOv2~\cite{dinov2}-based instance descriptors and training a SVM classifier in a single- or few-shot manner (depending on the scenario). We ablate on the descriptor used in the baseline method and show that among three popular state-of-the-art vision models (DINOv2, SAM and CLIP), DINOv2 embeddings are best at encoding instance knowledge.

Future work might focus on better leveraging motion and correlation in the video sequence. Another avenue would be building up richer, perhaps 3D, representations on-the-fly of the objects and using those for detection, tracking, segmentation and outlier filtering. Another possibility would be the use of a representation not directly based on point features, for example by modeling local neighborhoods or patches of features on the objects and using those to build a better, less ambiguous representation. Our goal is to build more malleable and teachable spatial AI systems and we hope to apply these techniques in such downstream systems. 
\section{Acknowledgements}

This work has received funding from the European Union’s Horizon 2020 research and innovation program under grant agreement No. 101017008 (Harmony).

{\small
\bibliographystyle{ieee_fullname}
\bibliography{egbib}
}

\end{document}